\documentclass[runningheads]{llncs}
\usepackage[T1]{fontenc}
\usepackage{amsmath}
\usepackage{graphicx}
\usepackage{subcaption}
\usepackage{makecell}
\usepackage{marvosym} 
\usepackage{amssymb}
\usepackage{multirow}
\usepackage{array}
\usepackage{hyperref}

\usepackage{booktabs}
\usepackage{upgreek}

\usepackage{verbatim}

\newcommand{\mypar}[1]{\vspace*{.3em}\noindent\textbf{#1}~}

\newcommand{\loss}{\mathcal{L}}

\newcommand{\mr}[1]{\mathit{#1}}

\newcommand{\ttimes}{\,$\times$\,}
\usepackage{xcolor}

\newcolumntype{C}[1]{>{\centering\let\newline\\\arraybackslash\hspace{0pt}}m{#1}}

\newsavebox\CBox

\newcommand{\ppm}{\,\tiny$\pm$}

\usepackage{xcolor}   
\usepackage{colortbl}
\newcommand{\ccol}{\cellcolor{gray!15}}

\begin{document}
%
\title{Automating MedSAM by Learning Prompts\\with Weak Few-Shot Supervision}
%
\titlerunning{Automating MedSAM with Weak Few-Shot Supervision}

%
\author{Mélanie Gaillochet\inst{1,2,3}\and
Christian Desrosiers\inst{1}\and
Hervé Lombaert\inst{1,2,3}
\authorrunning{M. Gaillochet et al.}
%
\institute{ÉTS Montréal, Canada 
\and
Polytechnique Montréal, Canada
\and
Mila - Quebec AI Institute, Université de Montréal, Canada}}
\maketitle              
\begin{abstract}
Foundation models such as the recently introduced Segment Anything Model (SAM) have achieved remarkable results in image segmentation tasks. However, these models typically require user interaction through handcrafted prompts such as bounding boxes, which limits their deployment to downstream tasks. Adapting these models to a specific task with fully labeled data also demands expensive prior user interaction to obtain ground-truth annotations. This work proposes to replace conditioning on input prompts with a lightweight module that directly learns a prompt embedding from the image embedding, both of which are subsequently used by the foundation model to output a segmentation mask. Our foundation models with learnable prompts can automatically segment any specific region by 1) modifying the input through a prompt embedding predicted by a simple module, and 2) using weak labels (tight bounding boxes) and few-shot supervision (10 samples). Our approach is validated on MedSAM, a version of SAM fine-tuned for medical images, with results on three medical datasets in MR and ultrasound imaging. Our code is available on \href{https://github.com/Minimel/MedSAMWeakFewShotPromptAutomation}{https://github.com/Minimel/MedSAMWeakFewShotPromptAutomation}.

\end{abstract}
\section{Introduction}
Annotation is a well-known labour-intensive and time-consuming task in medical imaging. Supervised segmentation models trained to identify specific regions of interest do not generalize well to new domains or classes and require more data and retraining when considering a new task. This increases the cost of developing segmentation models to solve multiple tasks. The need for universal models that can be applied to various tasks after training has hence been growing in medical image analysis. The introduction of foundation models for image segmentation such as the recent Segment Anything Model (SAM) \cite{kirillov_segment_2023}, as well as its versions adapted for medical imaging \cite{zhang_segment_2023}, notably MedSAM \cite{ma_segment_2024}, have appeared as a game-changer in the field of computer vision and medical image analysis. These models have shown remarkable performance on a variety of segmentation tasks. However, they remain promptable models that require user interaction to obtain the segmentation mask of a target object. Furthermore, their zero-shot performance depends on the quality of the user prompt. This reliance on user interaction hinders their integration into automatic pipelines and limits their usability at a large scale.

Recent attempts have been made to automate the prompt generation of SAM \cite{wu_self-prompting_2023,shaharabany_autosam_2023,zhang_personalize_2024}. However, these methods typically require samples with ground-truth segmentation masks, which are costly to obtain in the medical domain.

This paper proposes a lightweight add-on prompt module which learns to generate prompt embeddings directly from SAM's image embedding. Our end-to-end approach enables SAM models to specialize on the segmentation of a specific region and only requires few weakly-annotated samples. This reduces the interaction cost of developing specialized segmentation models. Our validation shows that, given only few training samples weakly annotated with tight boxes, promptable foundation model can effectively generate segmentation masks of target regions without requiring manual prompt inputs. 

\mypar{Foundation models for medical image segmentation.} Vision foundation models have achieved tremendous success in computer vision tasks thanks to large-scale pre-training. In particular, the Segment Anything Model (SAM) \cite{kirillov_segment_2023}, based on vision transformers \cite{dosovitskiy_image_2021} and trained on 1B masks and 11M images, was recently introduced as a prompt-driven foundation model for segmentation. 
Trained on natural images, SAM obtains uneven performances on medical data \cite{huang_segment_2024,mazurowski_segment_2023,wald_sammd_2023}, inducing its adaptation to the medical domain \cite{cheng_sam-med2d_2023,ma_segment_2024,shaharabany_autosam_2023}. In particular, MedSAM \cite{ma_segment_2024}, a foundation model for universal medical image segmentation was trained on 1.5 million image-mask pairs over 10 imaging modalities. These models provide impressive zero-shot performance, but remain promptable models that require user interaction at inference.

\mypar{Prompt automation for SAM.} 
Motivated by its performance in Natural Language Processing \cite{brown_language_2020}, prompt-tuning has successfully been applied to large vision models \cite{avidan_visual_2022}. Hence, methods that have focused on specializing SAM, a promptable model, have naturally explored prompt generation.
Given few fully labeled samples, the self-prompting unit of \cite{wu_self-prompting_2023} automatically generates a real point and bounding box from SAM's image embedding. AutoSAM replaces SAM's prompt encoder with a Harmonic Dense Net to adapt segmentation to medical images \cite{shaharabany_autosam_2023}.
A recent training-free approach, PerSAM \cite{zhang_personalize_2024}, encodes positive-negative location priors as prompt tokens to produces automatic segmentations of a specific object from a single reference image and mask. 
As opposed to our approach, all of these methods require samples with full segmentation masks.

\mypar{Segmentation with bounding box annotations.} Bounding boxes have emerged as an alternative to onerous annotation masks. Most methods use bounding boxes as an initial pseudo-label of the target region. A classic iterative graph-cut-based algorithm, GrabCut \cite{rother_grabcut_2004}, separates the foreground from its background given a bounding box. DeepCut \cite{rajchl_deepcut_2017} extends GrabCut to neural networks using existing heuristics. More recently, the bounding box tightness prior was adapted to deep learning-based models by imposing a set of constraints on the predictions \cite{kervadec_bounding_2020}, and was combined with multiple instance learning and smooth maximum approximation \cite{wang_bounding_2021}.\\

\mypar{Our contribution.} This work aims to efficiently automate MedSAM, a variant of SAM for the medical domain, to segment any target region through the use of few, weakly-labeled samples. Our approach introduces an innovative improvement by substituting the original prompt encoder, which requires user input, with an enhanced lightweight adaptable prompt-learning module that:
\begin{enumerate}\setlength\itemsep{.25em}
    \item Automatically \textbf{generates a prompt embedding} from the input image
    \item Trains with only \textbf{weak labels} (tight bounding boxes) and \textbf{few-shot} learning
    \item Is easily added on top of MedSAM (no fine-tuning)
\end{enumerate}

The next sections present our proposed prompt module for MedSAM and demonstrate its usefulness on various medical image segmentation tasks. 

\vspace*{-3pt}
\section{Methodology}
\vspace*{-3pt}

\vspace*{-3pt}
\subsection{Preliminaries: MedSAM architecture} 
\vspace*{-3pt}
Our approach builds upon on MedSAM \cite{ma_segment_2024}, a variant of SAM \cite{kirillov_segment_2023} fine-tuned on medical data. The model has three main components: a large image encoder $\mr{E}_{img}$, a prompt encoder $\mr{E}_{pr}$ and a lightweight mask decoder $\mr{D}_{mask}$. 

While the image encoder computes an embedding of the input image $x$, the prompt encoder outputs two sparse and dense embeddings from the provided set of prompts $[pr]$, respectively points or bounding boxes (BB), and a mask. The network produces a probability map $f_\theta$ by taking $x$ and a prompt embedding $Z_{pr} = \mr{E}_{pr}([pr])$:
$$f_\theta = \sigma \big(\mr{D}_{mask}(\mr{E}_{img}(x),\, Z_{pr})\big),$$
where $\sigma$ is the sigmoid function.

We present an end-to-end approach to remove the typical dependence on user-defined prompts $[pr]$, without modifying the pretrained MedSAM network.

\begin{figure}[ht]
    \centering
    \begin{subfigure}[b]{0.85\textwidth} 
        \centering
        \includegraphics[width=\textwidth]{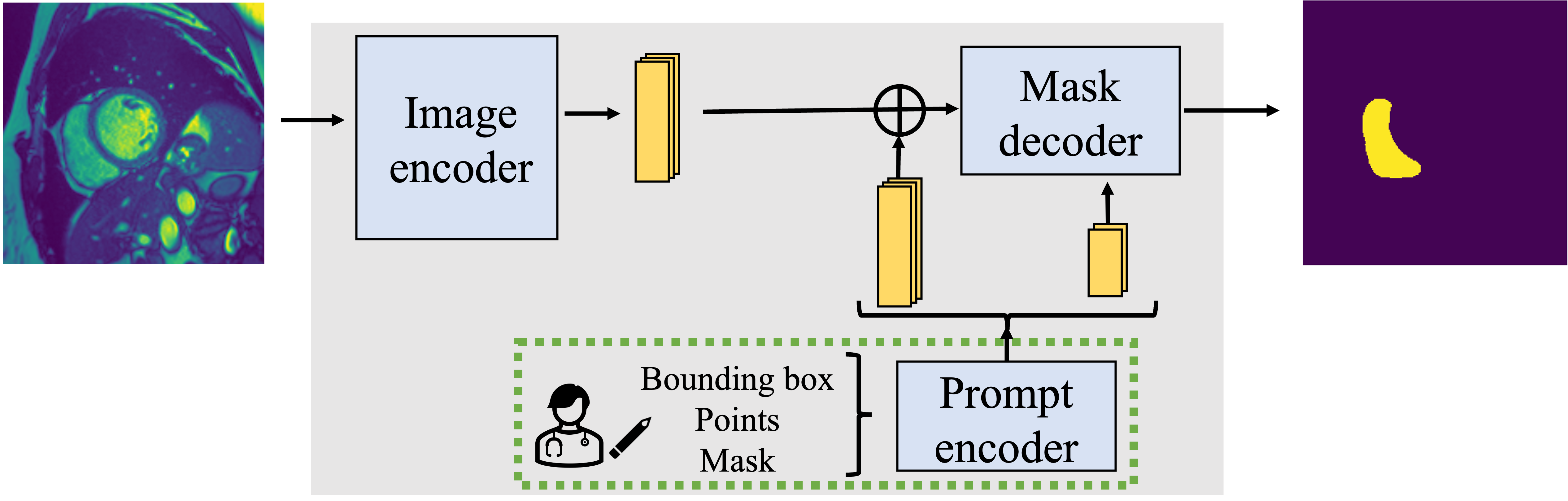}
        \caption{Promptable MedSAM \cite{ma_segment_2024}}
        \label{subfig:sam}
    \end{subfigure}
    \hfill 
    \begin{subfigure}[b]{0.80\textwidth} 
        \centering
        \includegraphics[width=\textwidth]{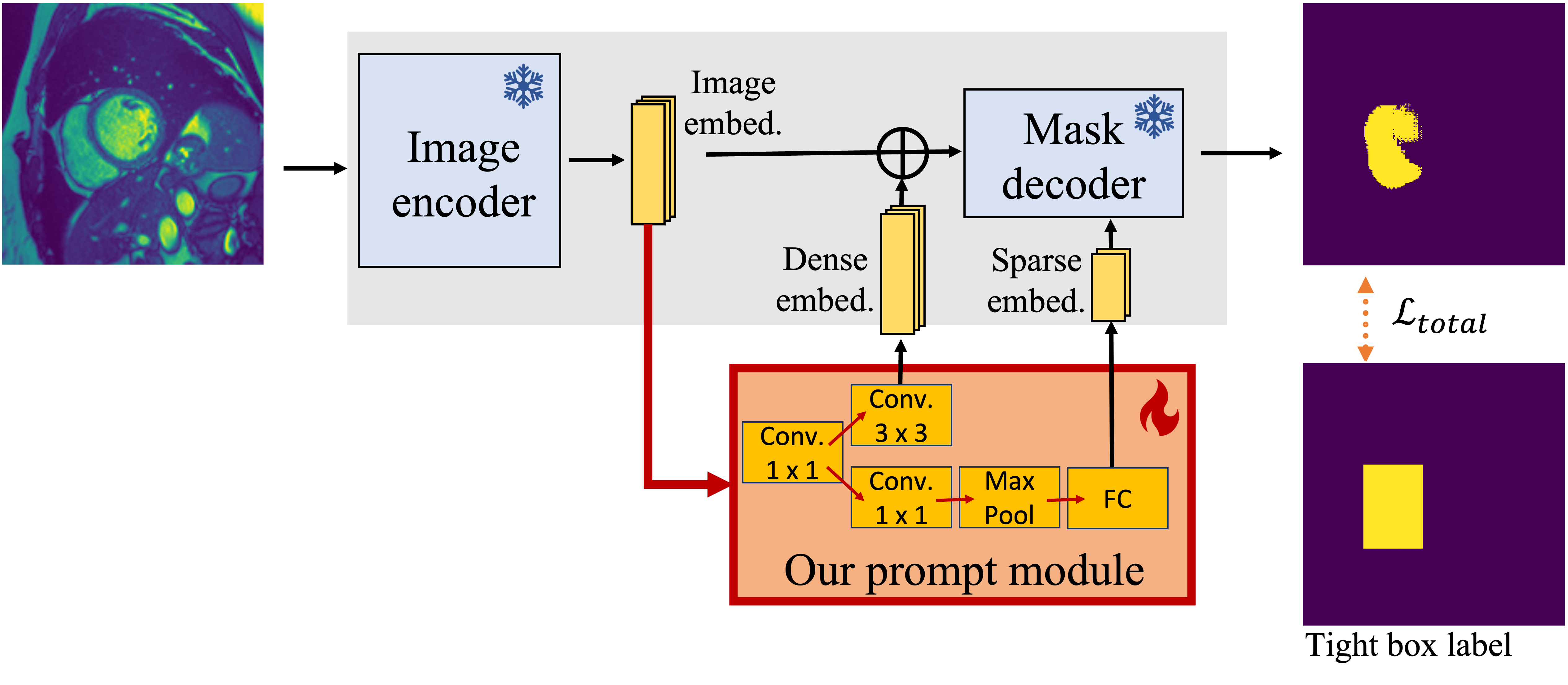}
        \caption{Automatic MedSAM (ours)}
        \label{subfig:method}
    \end{subfigure}
    \caption{Comparison between (\subref{subfig:sam}) MedSAM and (\subref{subfig:method}) our automation of MedSAM via a learnt prompt module. Our prompt module replaces MedSAM's prompt encoder and learns to generate a relevant prompt embedding from the image embedding. Training employs losses that utilize only tight box labels.
    }
    \vspace*{-3pt}
    \label{fig:method}
\end{figure}

\vspace*{-3pt}
\subsection{Lightweight prompt module}
\vspace*{-3pt}
Our approach consists of a prompt module trained to compute directly $Z_{pr}$ from the image embedding provided by MedSAM (see Fig.\ref{subfig:method}). The module outputs two embeddings of the same shape as those generated by MedSAM (Fig.\ref{subfig:sam}). Originally, the dense prompt embedding has a spatial correspondence with the image and can be considered as a low-quality segmentation map, while the sparse embeddings are spatial encodings of coordinates. Therefore, our prompt module generates a dense embedding through a convolutional layer and a sparse embedding through a fully connected (FC) layer. 

\vspace*{-3pt}
\subsection{Learning with tight box annotations}
\vspace*{-3pt}
Denote as $X: \Upomega \subset \mathbb{R}^{3\times H \times W} \rightarrow \mathbb{R}$ a 3-channel input image of height $H$ and width $W$, where $\Upomega$ is the spatial domain corresponding to each channel of the image. Moreover, let $Y \in \{0,1\}^{\Upomega}$ be the ground-truth binary segmentation mask of $X$. Suppose we only have access to a tight bounding box $\tilde Y$ of the target. $\Upomega_I$ and $\Upomega_O$ define the regions respectively inside and outside the bounding box such that  $\Upomega_I + \Upomega_O = \Upomega$. This leads to a constrained optimization problem \cite{kervadec_bounding_2020} from the bounding box annotations $\tilde Y$.

\mypar{Emptiness of $\Upomega_O$.} Since the region in $\Upomega_I$ defined by the bounding box must contain the target object, $\Upomega_O$ must contain only foreground. Hence, we can apply a Cross-entropy loss for all pixels $p \in \Upomega_O$: 
\begin{equation}
    \label{eq:emptiness_loss}
    \loss_{\mr{empty}} \, = \, -\sum_{p \in \Upomega_O} \!\log (1 - f_\theta (p)).
\end{equation}

\mypar{Tight box constraint in $\Upomega_I$.} The tightness of the bounding box indicates that at least one foreground pixel must cross every horizontal and vertical line of weak label $\tilde Y$. As in \cite{kervadec_bounding_2020}, we soften this condition by considering segments of width $w$ instead of individual lines and ensure differentiability by considering output probabilities instead of the prediction mask. The condition formalizes as:
\begin{equation}
    \label{eq:tightbox_constraint}
    \sum_{p \in s_l} f_\theta (p) \geq w, \ \ \forall s_l \in S_L,
\end{equation}
where $S_L$ is the set of all vertical and horizontal segments of width $w$ that make the bounding box $\tilde Y$. We convert the inequality constraints of \eqref{eq:tightbox_constraint} to a loss using a penalty function $\psi_t$, and obtain:
\begin{equation} \label{eq:tightbox_loss}
\loss_{\mr{tightbox}} \, = \, \sum_{s_l \in S_L} \psi_t \bigg (w -  \sum_{p \in s_l} f_\theta (p)\bigg).
\end{equation}
The penalty function can be modeled as a simple scaled ReLU function, i.e. $\psi_t(x) = t\!\cdot\!\max(0, x)$. In this work, we instead resort to a pseudo log-barrier function, which provides a more stable optimization under multiple competing constraints. As $t\!\to\!\infty$, function $\psi_t(x)$ behaves as a hard barrier where $\psi_t(x)=\infty$ if $x > 0$, else $\psi_t(x)=0$. In our method, we found that using a fixed value of $t=5$ worked best.

\mypar{Foreground size constraint.} The bounding box $\tilde Y$ also sets a limit on the target size of the prediction mask. Again, we consider output probabilities rather than individual predictions to ensure differentiability. By applying priors on the fraction $\epsilon \in [0, 1]$ of pixels from $\Upomega_I$ that belong to the background, we get:
\begin{equation}
    \label{eq:size_constraint}
    \epsilon_1 |\Upomega_I| \leq \sum_{p \in \Upomega} f_\theta (p) \leq \epsilon_2|\Upomega_I|.
\end{equation}
As before, we employ $\psi_t(x)$ to convert these inequality constraints into the following loss:
\begin{equation} \label{eq:size_loss}
    \loss_{\mr{size}} \, = \, \psi_t \bigg(\epsilon_1 |\Upomega_I| - \sum_{p \in \Upomega} f_\theta (p))\bigg) + \psi_t \bigg(\sum_{p \in \Upomega} f_\theta (p) - \epsilon_2 |\Upomega_I|\bigg).
\end{equation}

Given \eqref{eq:emptiness_loss}, \eqref{eq:tightbox_loss} and \eqref{eq:size_loss}, and weights $\lambda_1$ and $\lambda_2$, the final loss becomes:
\begin{equation}
\loss_{\mr{total}} \, = \,\loss_{\mr{empty}}
     \, + \, \lambda_1\,\loss_{\mr{tightbox}}  \, +\,  \lambda_2\,\loss_{\mr{size}}.
\end{equation}

\section{Results}
\vspace*{-3pt}
\subsection{Datasets} \label{subsec:data} 
\vspace*{-3pt}
Our experiments validate our method on three public datasets: the Head Circumference dataset\footnote{https://hc18.grand-challenge.org/} (HC18) \cite{heuvel_automated_2018}, the Cardiac Acquisitions for Multi-structure Ultrasound Segmentation\footnote{https://www.creatis.insa-lyon.fr/Challenge/camus/} (CAMUS) \cite{leclerc_deep_2019} and the Automated Cardiac Diagnosis Challenge\footnote{https://humanheart-project.creatis.insa-lyon.fr/database/} (ACDC)~\cite{bernard_deep_2018}. For both cardiac datasets, the end diastole images are used.
For HC18, we filter out samples with ground-truth masks that could not be automatically generated by OpenCV from the circumference annotations, and split the ultrasound dataset into 507 training, 77 validation and 148 test images.
For CAMUS, we focus on the left ventricle (LV) and left atrium (LA) segmentation and use 50 images for validation, 100 images for testing and the remaining 350 images for training. 
For ACDC, we focus on the right ventricle (RV) and LV segmentation and use 10 patients for validation (78 images), 50 for testing (470 images)and the remaining 90 patients (765 images) for training. Each sample has a minimum foreground size for all experiments.

Following \cite{ma_segment_2024}, our preprocessing includes clipping the intensity values of each 2D image (HC18, CAMUS) or each 3D volume (ACDC) between the $0.5^{th}$ and $99.5^{th}$ percentiles and rescaling them to the range [0, 255]. We also partition each 3D volume of ACDC into 2D image which we resample to a fixed 1mm\ttimes1mm resolution. We center crop and pad each sample to size 640\ttimes640 (HC), 512\ttimes512 (CAMUS) or 256\ttimes256 (ACDC). To meet MedSAM's requirements, we resize all images to a fixed 3\ttimes1024\ttimes1024 size before inputting them in the model.

\vspace*{-3pt}
\subsection{Implementation details}
\vspace*{-3pt}

\mypar{Model.} Our backbone model is MedSAM based on ViT-B, the smallest version of SAM. The backbone remains frozen during training. We keep our prompt module lightweight by using few layers. A 1\ttimes1 convolution first reduces the number of channels. Then, the dense embedding is obtained through a 3\ttimes3 convolution, while the sparse embedding is obtained through a 1\ttimes1 convolution followed by max pooling and a fully connected layer. All convolutional layers are followed by ReLU activation. Our prompt module has thus 2.4M trainable parameters.

\mypar{Loss parameters.} We train our prompt module using $\loss_{\mr{total}}$, with $\lambda_1 = 0.0001$, $\lambda_2 = 0.01$. For our tight box constraint, we follow \cite{kervadec_bounding_2020} and use segments of $w=5$. We hypothesize that the foreground region is at least half the size its tight bounding box and set $[\epsilon_1, \epsilon_2] = [0.5, 0.9]$. Comparative training with full segmentation masks uses a Binary Cross-entropy Dice loss, each term having the same weight.

\mypar{Training.}
We use a batch size of 4 and a learning rate (LR) of $0.001$ with a multi-step scheduler decreasing LR by 0.1 after half the epochs and a weight decay of 0.0001. 
To minimize computational complexity, we do not use data augmentation. This allows us to discard MedSAM's image and prompt encoders after saving the image embeddings during an initial iteration, reducing the number of total parameters from 96.1M to only 6.5M (2.4M trainable). In the 10-shot setting, we repeat the experiments 9 times, with 3 initialization seeds and 3 training subsets selected uniformly at random. The results are averaged over these experiments. All experiments are implemented in Python 3.8.10 with Pytorch on NVIDIA RTX-A6000 GPUs.

\mypar{Baselines.}
For each class, we compare our method with two specialized single-task models: a standard UNet \cite{ronneberger_u-net_2015} and a TransUNet \cite{chen_transunet_2021}. We also validate our method against PerSAM \cite{zhang_personalize_2024} which automates SAM with 1-shot supervision, and AutoSAM \cite{shaharabany_autosam_2023} which uses a Harmonic Dense Net (41.6M parameters) to learn the prompt embedding. We train the UNet, TransUNet and AutoSAM on full segmentation masks with a standard Cross-entropy Dice loss. To improve the performance of the baseline models, longer training is used (200 epochs) with a larger batch size of 24 for TransUNet (following \cite{chen_transunet_2021}). Similarly, the best results for PerSAM are obtained with the ViT-H backbone.
Results for MedSAM are also included when prompted with the tightest bounding boxes (no noise).

\subsection{Validation on multiple medical segmentation tasks}

\begin{table}[t!]
\centering
\caption{Model performance on test sets in terms of mean (std) 2D Dice similarity score ($\uparrow$). Best results in few-shot settings are highlighted in bold.
}
\label{tab:modelPerformance}
\setlength{\tabcolsep}{4pt}
\resizebox{\linewidth}{!}
{
\begin{tabular}{l|l|c|cc|c|cc|cc}
\toprule
\multirow{2}{*}[-2pt]{Type} &  \multirow{2}{*}[-2pt]{Method} &  \multirow{2}{*}[-2pt]{\makecell[c]{\# \\ Samples}} & \multirow{2}{*}[-2pt]{\makecell[c]{Mask \\ labels}} &  \multirow{2}{*}[-2pt]{\makecell[c]{BB \\ labels}} &  \multirow{2}{*}[-2pt]{HC} & \multicolumn{2}{c|}{CAMUS} & \multicolumn{2}{c}{ACDC}\\
\cmidrule(l{4pt}r{4pt}){7-8}\cmidrule(l{4pt}r{4pt}){9-10}
 &    &  &  &  &  & LV & LA & RV & LV\\
\midrule\midrule
 Promptable & \makecell[l]{MedSAM \cite{ma_segment_2024} \\ (w/ tight box)} & -- &  --& --  & 95.19 & 94.50 & 89.23 & 93.78 & 95.45\\
 \midrule
  \multirow{4}{*}{\makecell[l]{Automatic \\ (fully trained)}} & \multirow{2}{*}{UNet \cite{ronneberger_u-net_2015}} & All & \checkmark &  & 86.53\ppm0.55 & 89.93\ppm0.01& 74.77\ppm0.78& 89.55\ppm0.23& 94.83\ppm0.13\\
  &   & 10 & \checkmark &  & 61.79\ppm3.10 & 75.09\ppm3.69 & 46.29\ppm3.64 & 40.85\ppm1.66 & 59.96\ppm0.91 \\
  \cmidrule(l{2pt}r{2pt}){2-10}
&    \multirow{2}{*}{TransUNet \cite{chen_transunet_2021}} & All & \checkmark &  &  96.32\ppm0.19 & 92.92\ppm0.29 & 85.04\ppm0.15 & 90.79\ppm0.07 & 94.08\ppm0.07\\
 &     & 10 & \checkmark &  & \textbf{92.15}\ppm0.40& 87.32\ppm0.45 & 66.51\ppm2.28 & \textbf{68.69}\ppm0.58 & 78.98\ppm1.36 \\
 \midrule
 \multirow{5}{*}{\makecell[l]{Automatic \\ (adapted)}} &  \multirow{2}{*}{AutoSAM \cite{shaharabany_autosam_2023}} & All & \checkmark &  & 97.42\ppm0.04 & 93.59\ppm0.03 & 85.60\ppm0.89 & 89.57\ppm0.54 & 95.18\ppm0.11 \\
 &  & 10 & \checkmark &  & 90.64\ppm1.84 & 86.98\ppm0.67 & 67.09\ppm4.82 & 68.33\ppm3.21 & \textbf{84.17}\ppm2.05\\
 \cmidrule(l{2pt}r{2pt}){2-10}
 &  PerSAM \cite{zhang_personalize_2024} & 1 & \checkmark &  &  58.98\ppm0.19& 36.13\ppm0.00 & 14.19\ppm0.02 &  27.64\ppm9.48 & 45.43\ppm5.47\\
\cmidrule(l{2pt}r{2pt}){2-10}
 & \ccol & \ccol All &  \ccol & \ccol \checkmark & \ccol 92.88\ppm1.27 & \ccol 88.86\ppm1.42& \ccol 79.82\ppm0.74& \ccol 76.97\ppm1.02& \ccol 86.91\ppm2.08\\
& \ccol \multirow{-2}{*}{Ours}  & \ccol 10 &  \ccol & \ccol \checkmark & \ccol 85.23\ppm0.55 & \ccol \textbf{88.38}\ppm0.83 & \ccol \textbf{73.56}\ppm0.57 & \ccol 58.96\ppm2.28 & \ccol 80.37\ppm1.59\\
\bottomrule
\end{tabular}
}
\end{table}

\begin{figure}[t!]
\centering
\setlength{\tabcolsep}{1pt}
\begin{tabular}{ccccccc}
    \includegraphics[width=.16\linewidth]{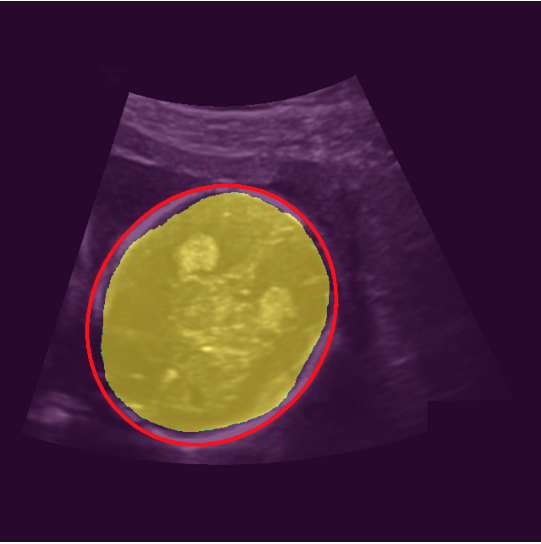} &
    \includegraphics[width=.16\linewidth]{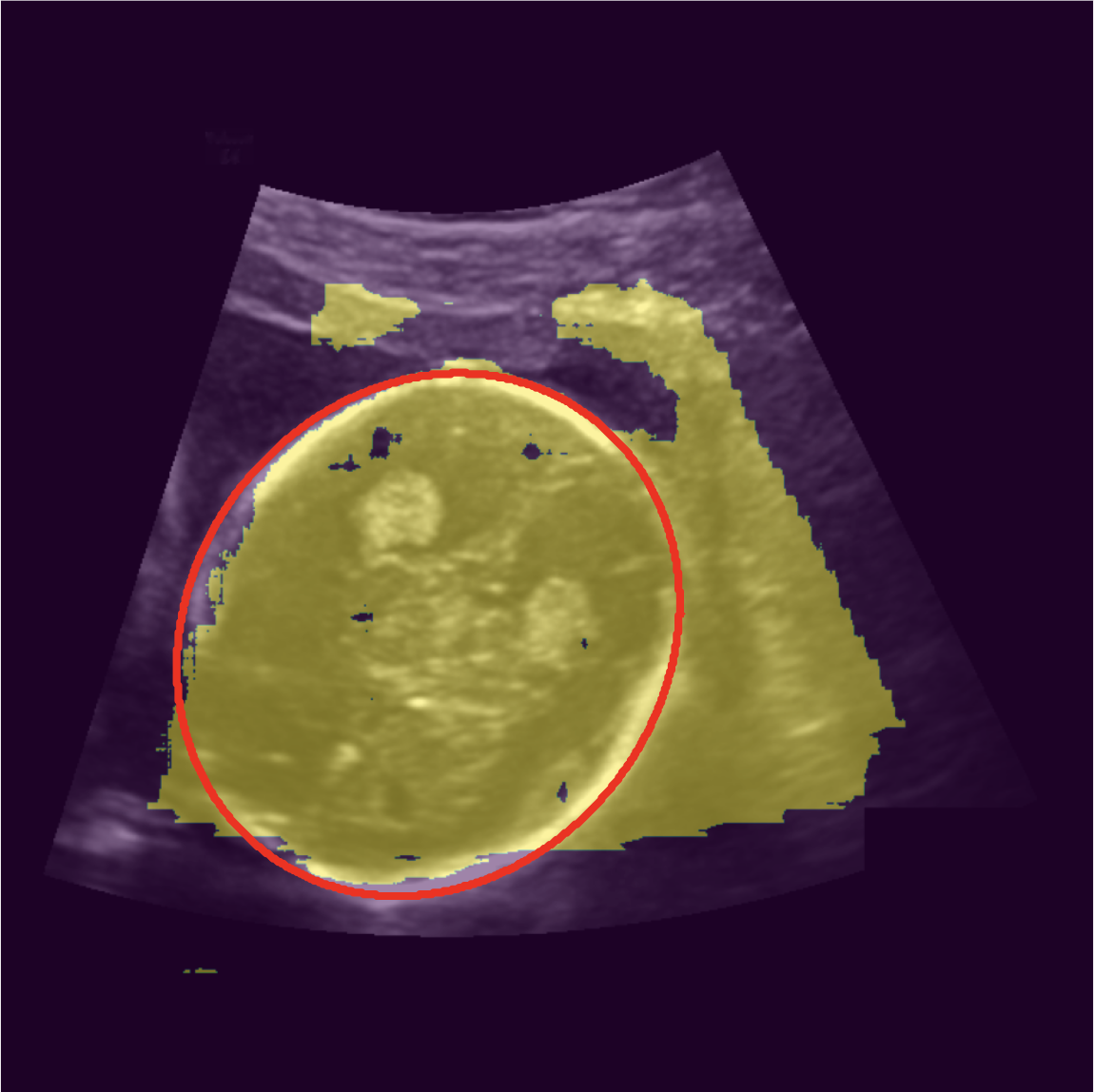} &
    \includegraphics[width=.16\linewidth]{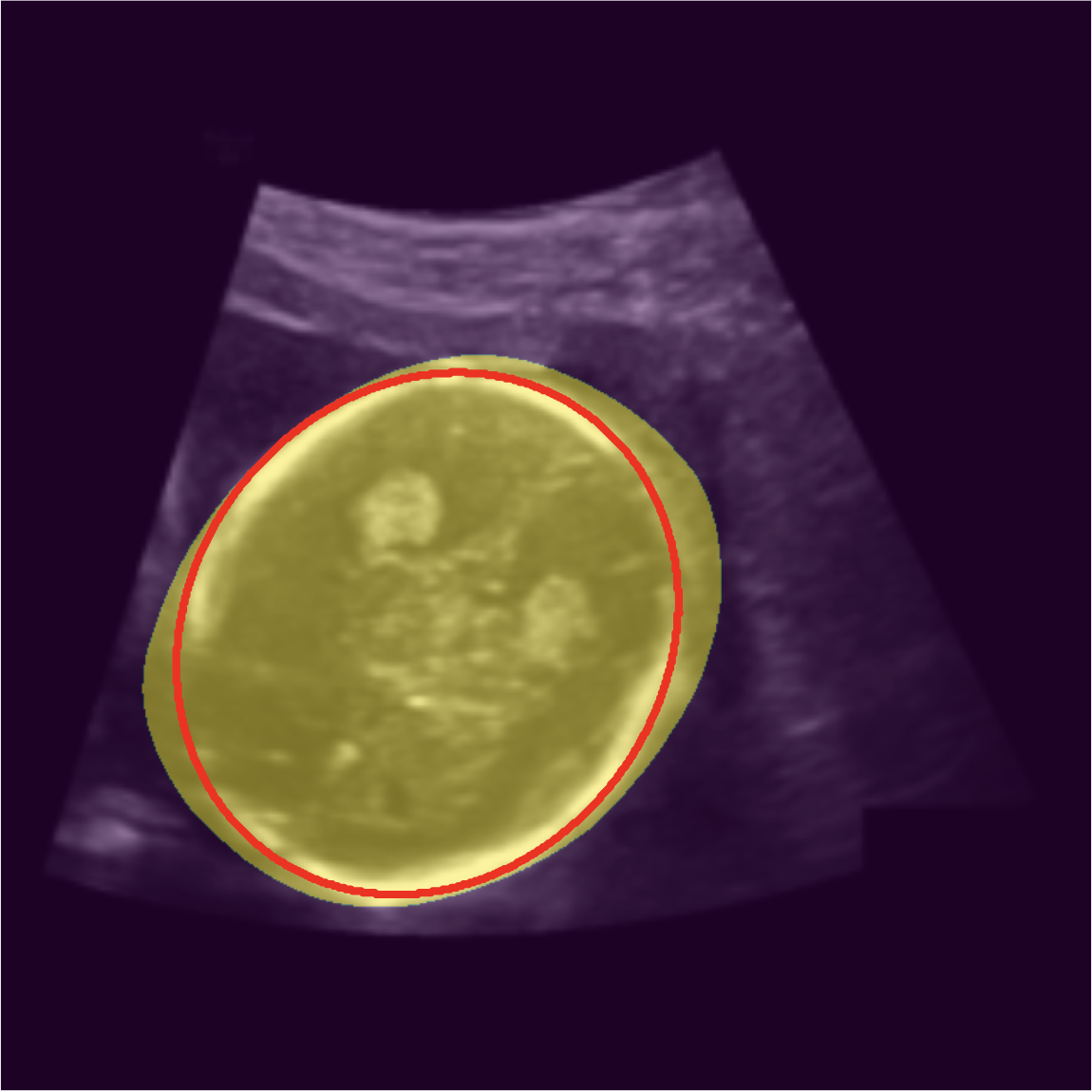} &
    \includegraphics[width=.16\linewidth]{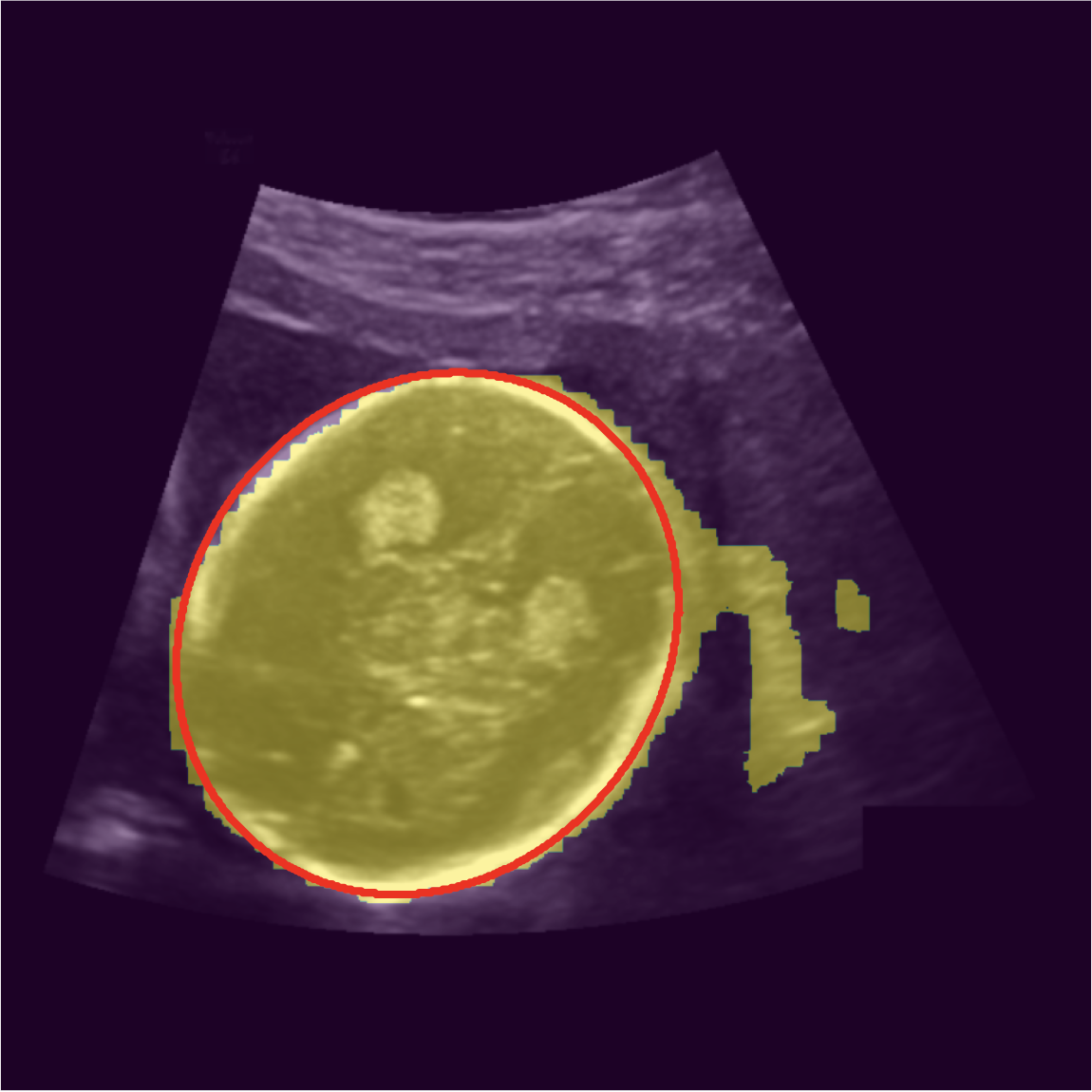} & 
    \includegraphics[width=.16\linewidth]{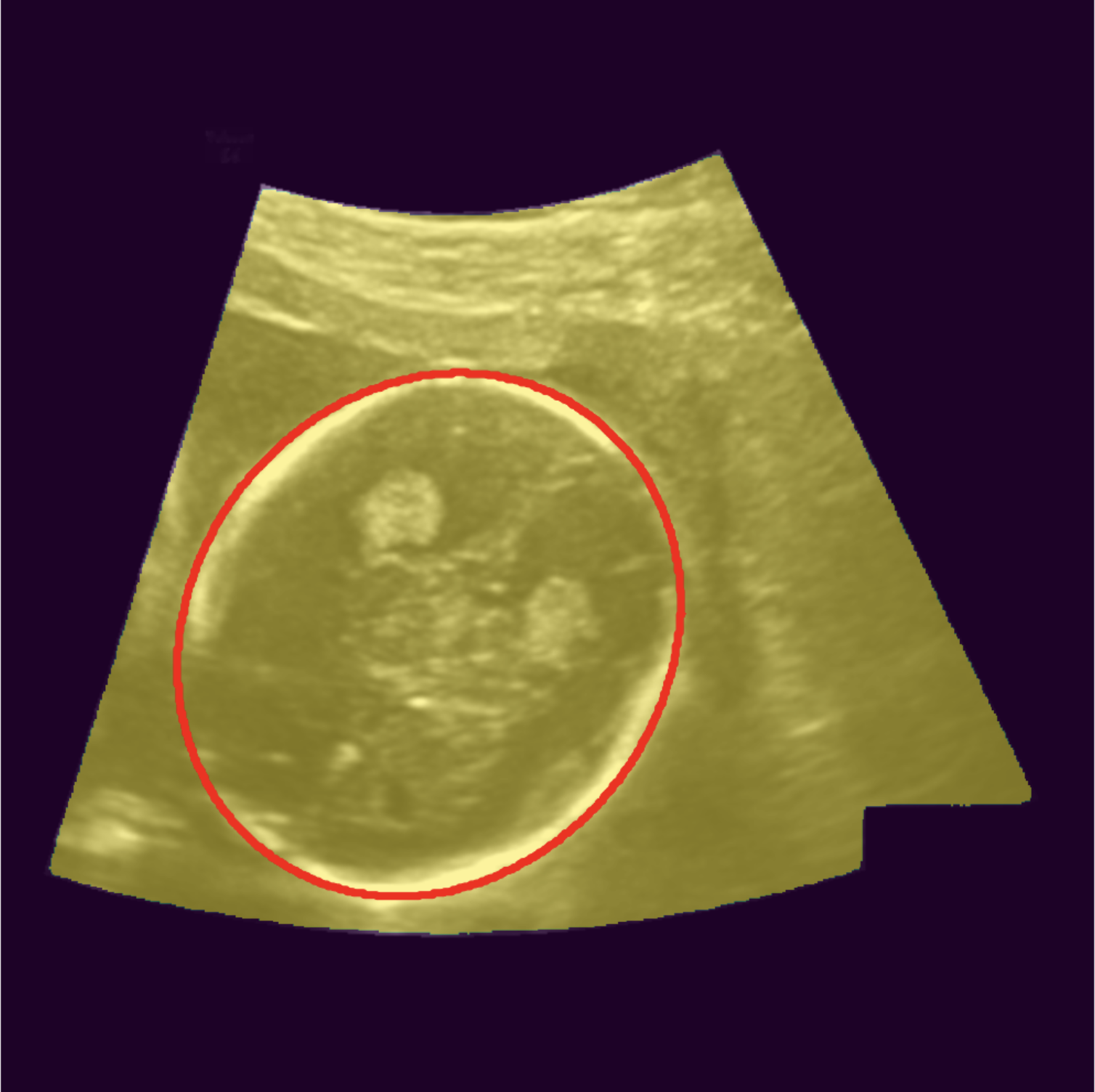} & 
    \includegraphics[width=.16\linewidth]{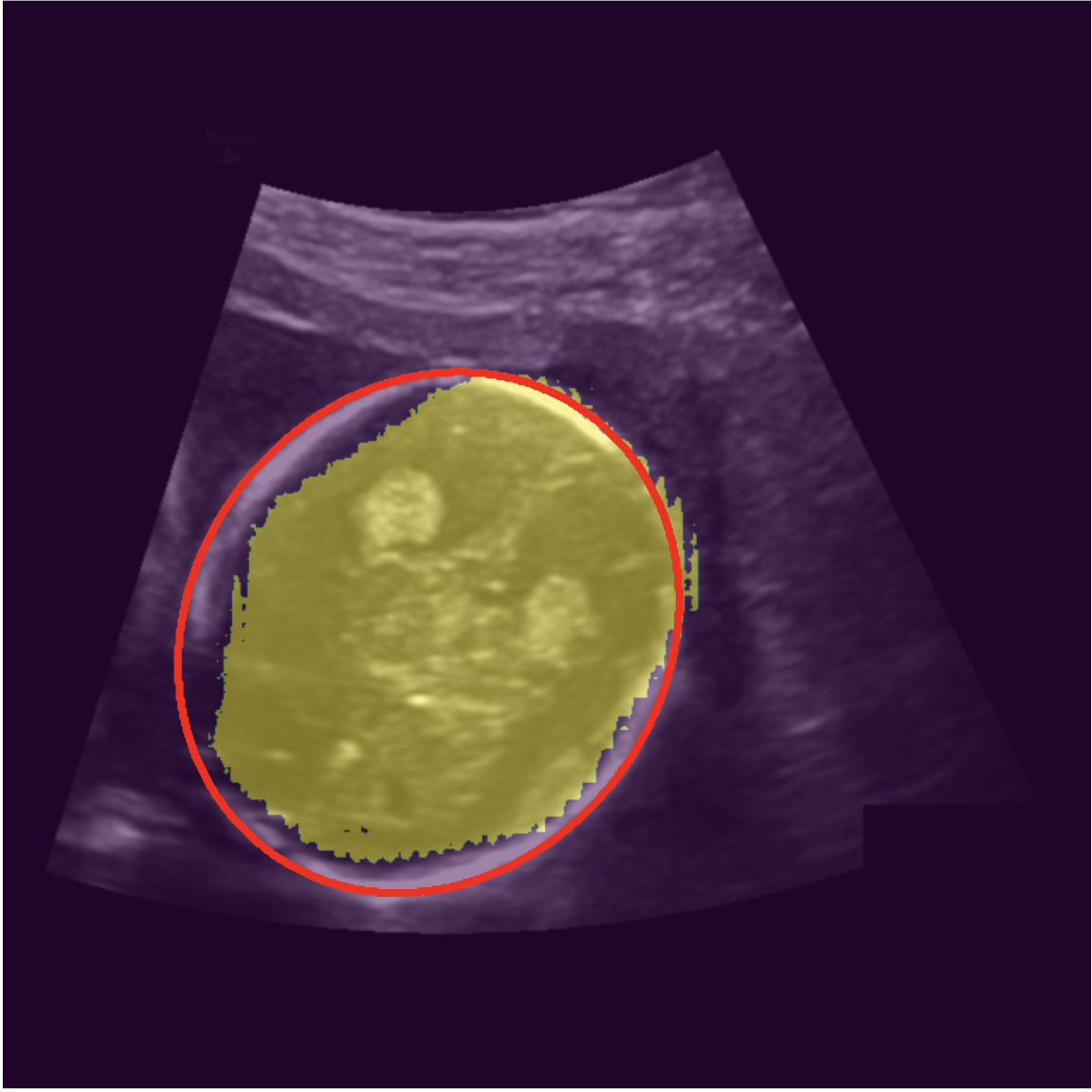} \\
    
    \includegraphics[width=.16\linewidth]{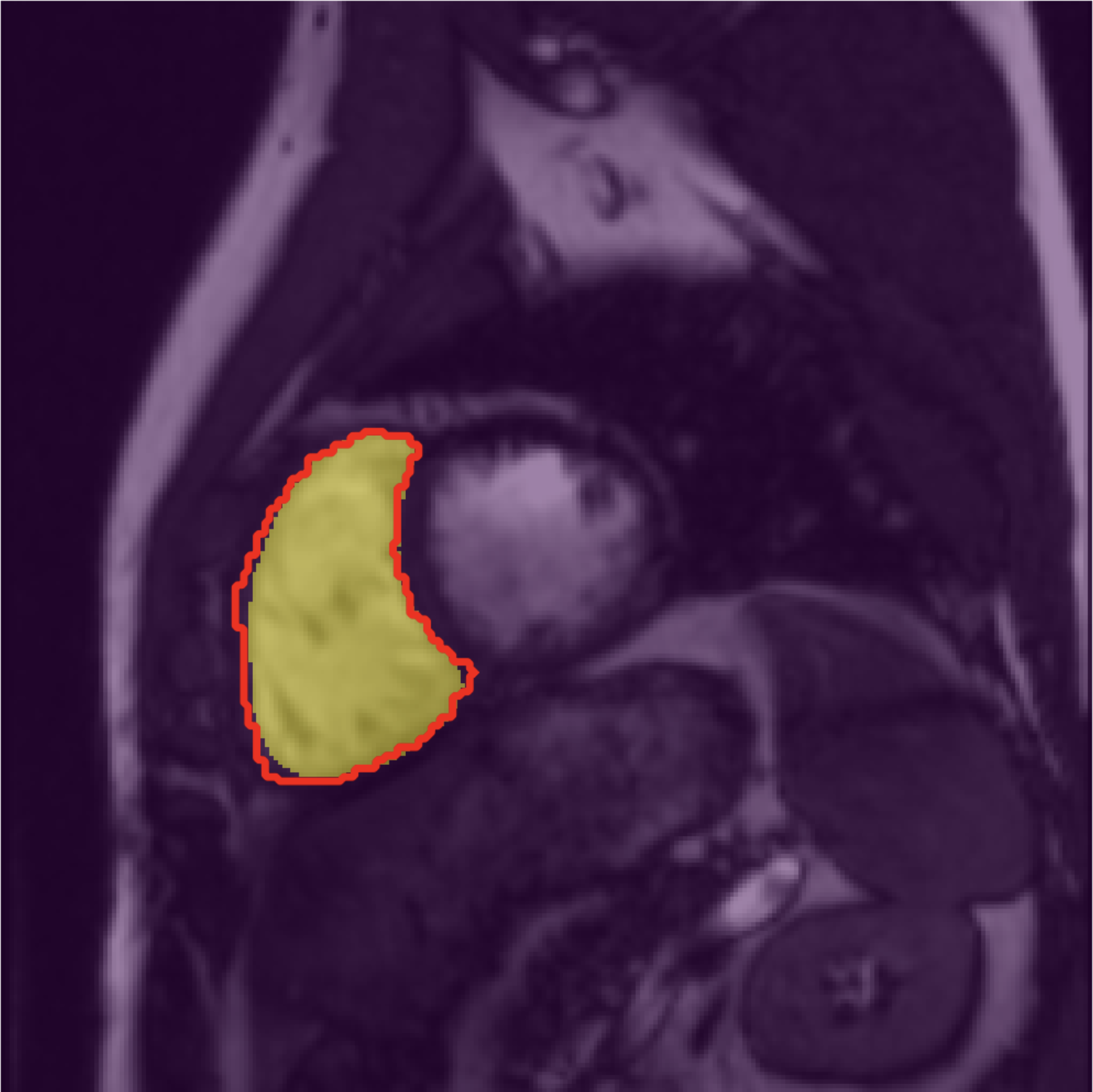} & 
    \includegraphics[width=.16\linewidth]{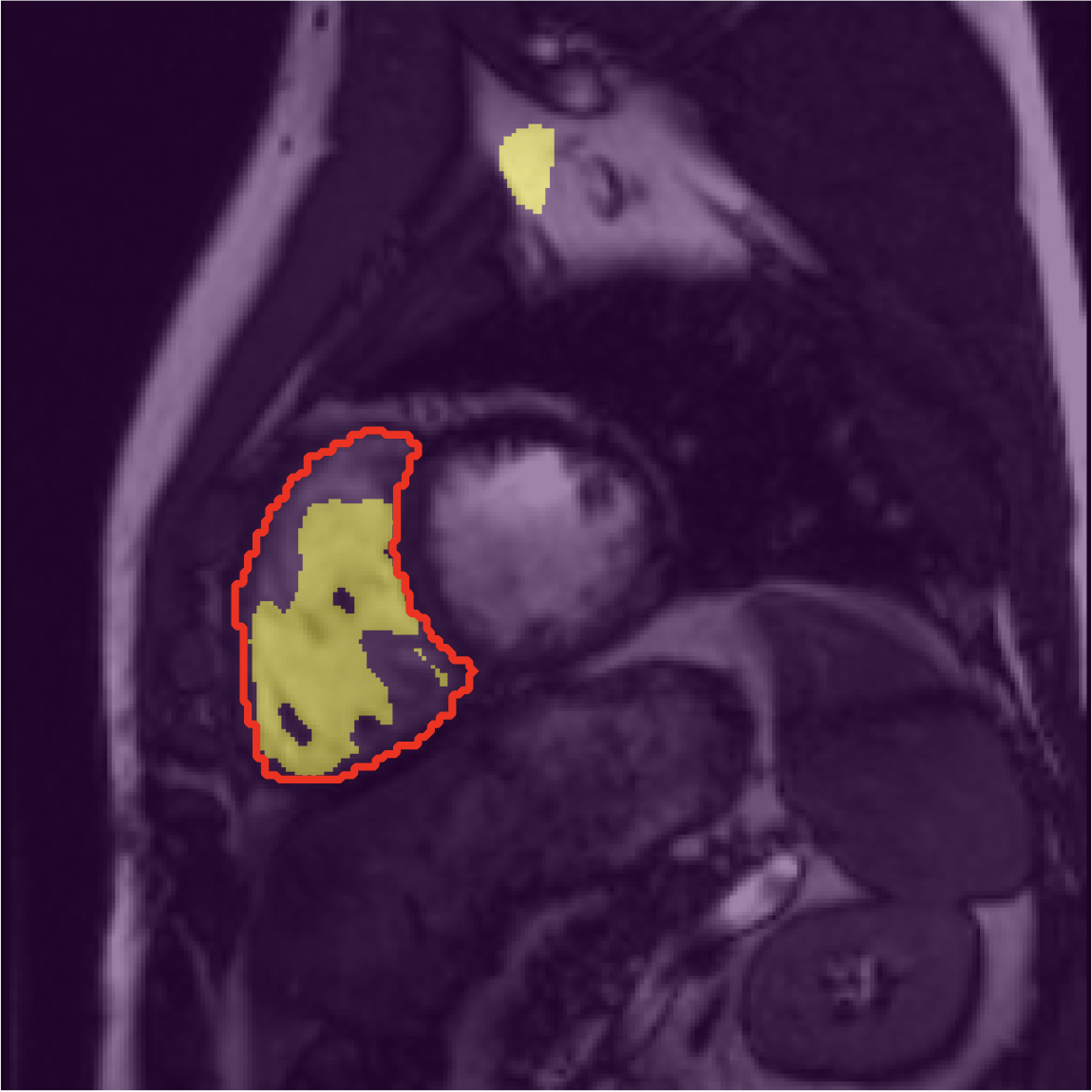} &
    \includegraphics[width=.16\linewidth]{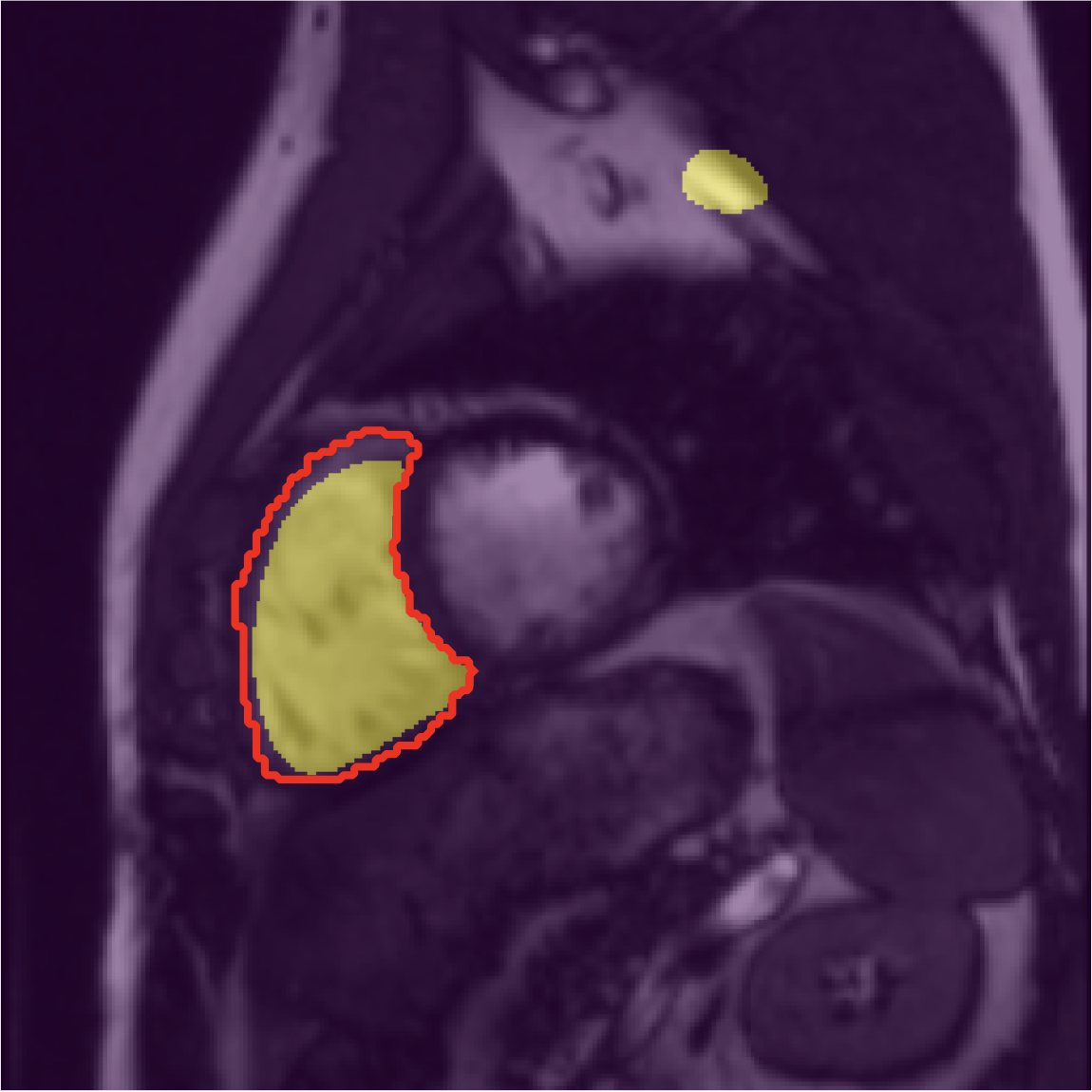}  &
    \includegraphics[width=.16\linewidth]{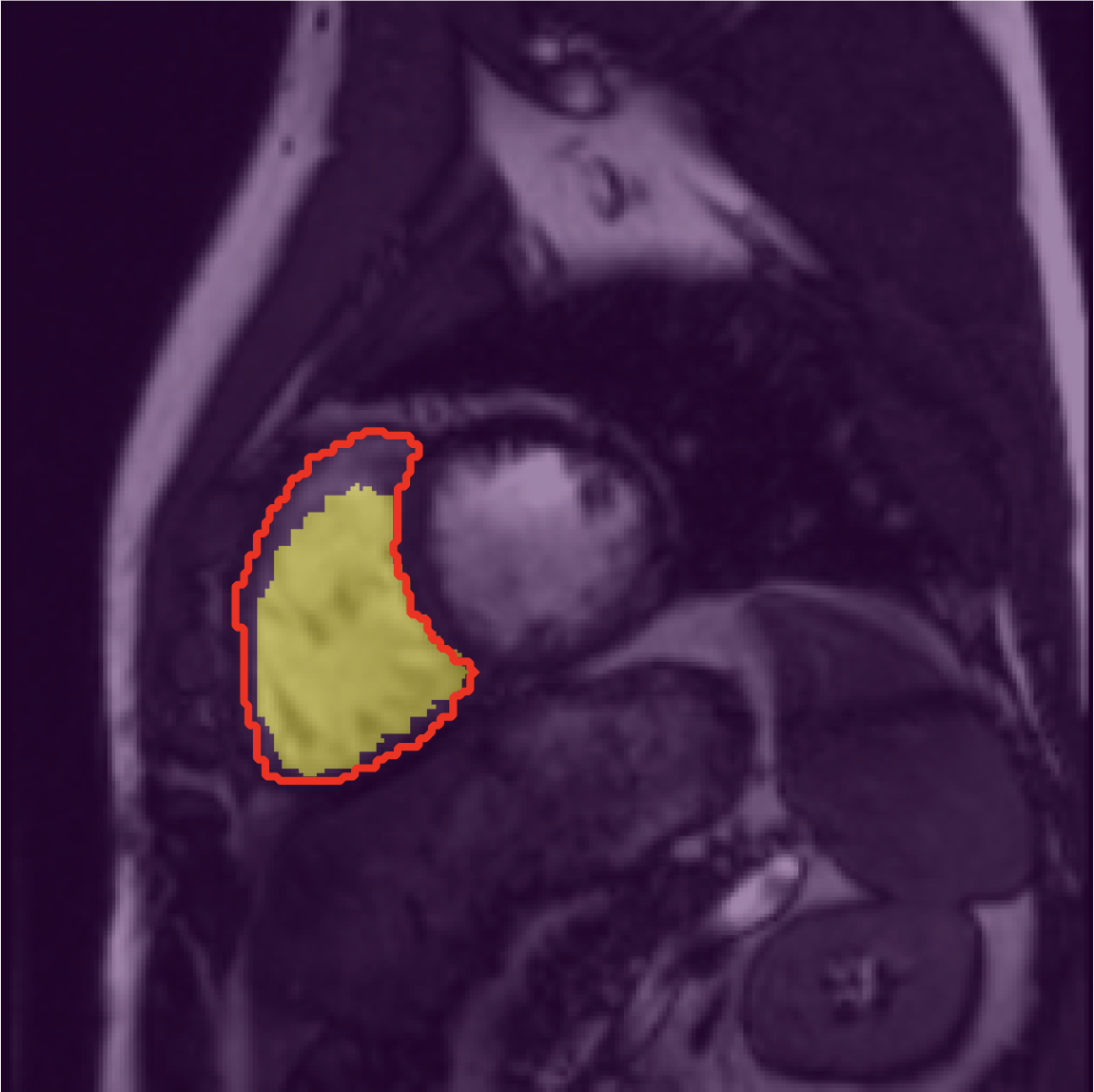} & 
    \includegraphics[width=.16\linewidth]{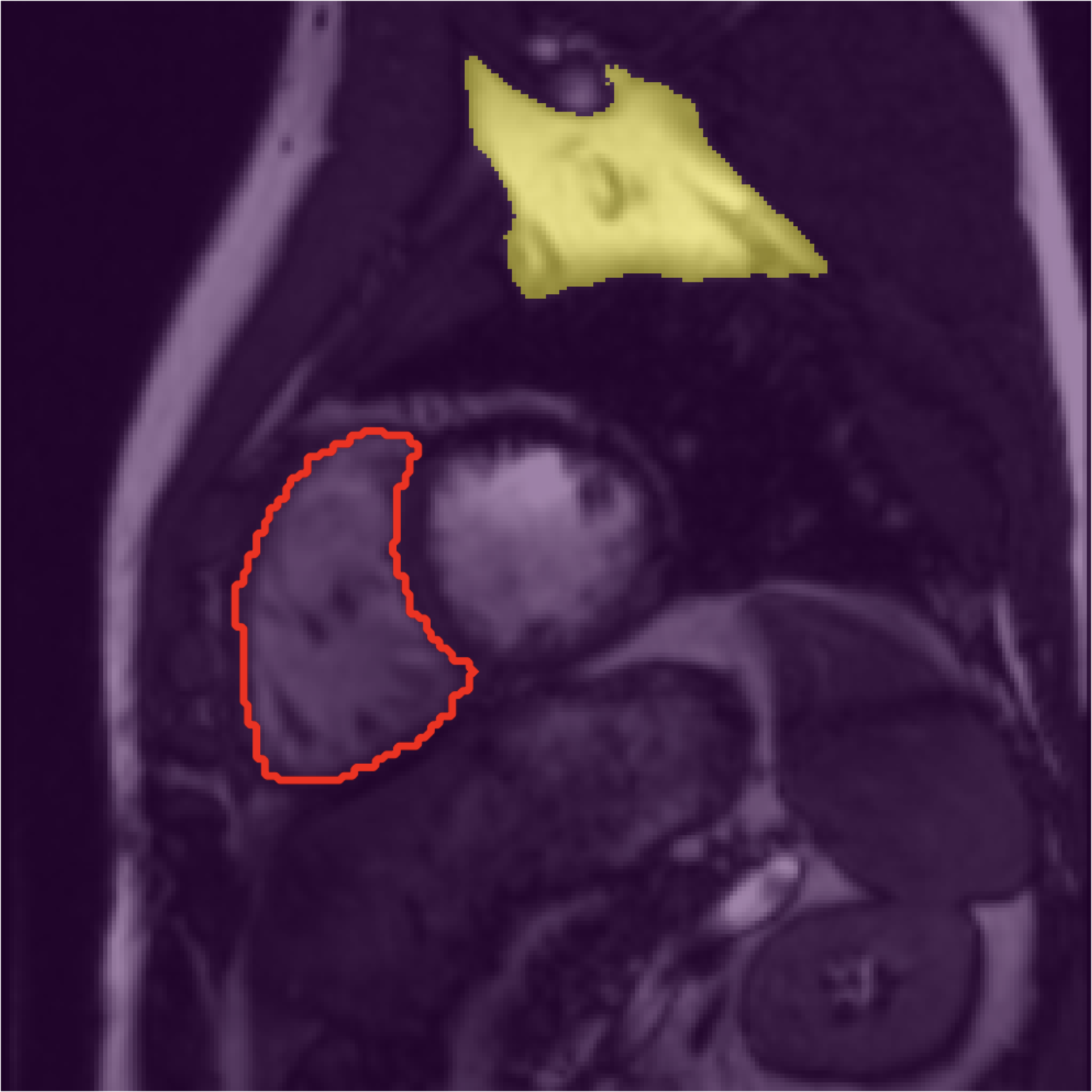} & 
    \includegraphics[width=.16\linewidth]{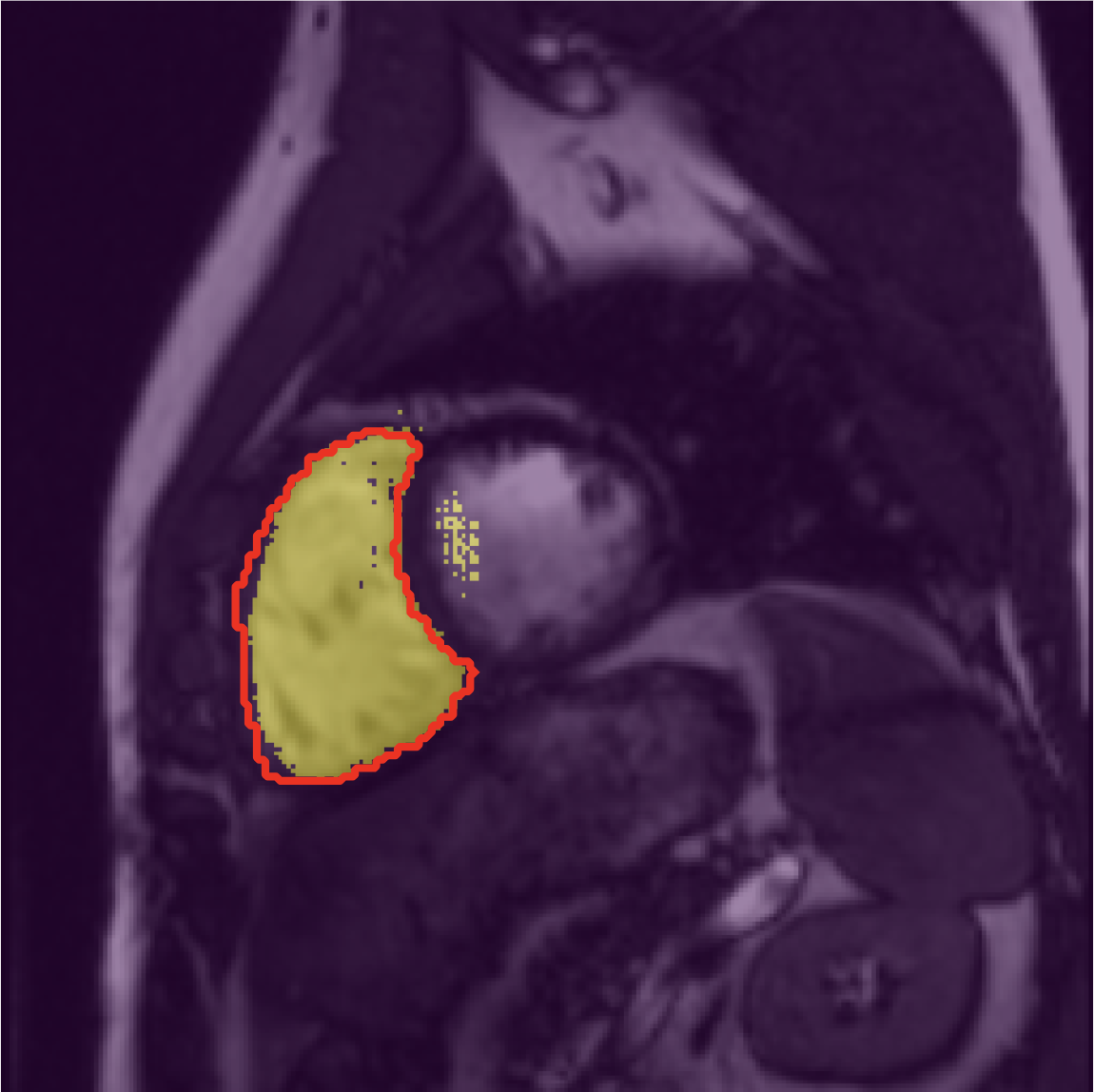} \\
      \makecell[c]{(a) MedSAM  \\ (Prompted)} &
      \makecell[c]{(b) UNet \\  (10 masks)} & 
      \makecell[c]{(c) TransUNet \\  (10 masks)} & 
      \makecell[c]{(d) AutoSAM \\  (10 masks)} & 
      \makecell[c]{(e) PerSAM \\  (1 mask)} & 
      \makecell[c]{(f) Ours \\ (10 BB)} & 
\end{tabular}
    \caption{Predicted segmentations on test samples of HC18 (row 1) and the right ventricle in ACDC (row 2). From left to right, (a) MedSAM prompted with a tight box, (b-d) UNet, TransUNet and AutoSAM, trained with ground-truth masks, (e) PerSAM using one reference image with its ground-truth mask, and (f) our method trained on tight bounding boxes. All automatic methods are given for the 10-shot setting, except PerSAM, a 1-shot approach. Ground-truth annotation is drawn in red, with predicted segmentation mask overlayed in yellow. 
    }
    \label{fig:examples}
    \vspace*{-1em}
\end{figure}

The Dice similarity score (DSC) is used as our evaluation criteria. We evaluate our approach on three datasets: CAMUS, an internal validation set of MedSAM, and HC18 and ACDC, two datasets never seen by MedSAM during training. This allows us to verify the ability of the prompt module to effectively learn which region to segment in both in-domain and out-of-domain data. Training is performed on both the entire training set and the difficult 10-shot regime.

Our results are given in Table \ref{tab:modelPerformance} and Fig.\ref{fig:examples}. First, with only tight bounding box (BB) annotations, our approach trained on all samples is able to outperform a UNet trained on ground-truth segmentation masks for 2 different tasks (HC and LA). 
The most significant results are observed in the 10-shot setting. With only $1.3\%$ (ACDC), $2\%$ (HC18) and $2.9\%$ (CAMUS) of the total training samples, our approach sees only a slight decrease in performance (except for RV segmentation) compared to the full-data setting. This contrasts with the considerable performance drop observed with UNet and TransUNet in multiple segmentation tasks, even when both methods are trained with ground-truth mask labels. Therefore, our prompt module-based approach is not only more computationally efficient to train than specialized models, but it also requires only weak labels and appears more robust in the few-shot setting.
AutoSAM displays slightly better test dice scores than our approach, but AutoSAM requires full ground-truth masks and uses a much heavier model to learn the prompt, yielding a 3-fold increase in the training time.
Additionally, PerSAM, which uses only one reference image, fails to generate convincing segmentations. Its poor performance on medical datasets may be due to the fact that PerSAM generates point prompts used by SAM, which are more likely to introduce ambiguity \cite{ma_segment_2024}.


The benefits of our proposed methods are visually supported by Fig.\ref{fig:examples}. Our module trained on 10 training samples and tight bounding boxes yields segmentations much more convincing than those produced by a UNet trained on ground-truth masks. Given little training data, the UNet hallucinates large regions in the background. Our approach is also able to generate segmentation masks more faithful to the ground-truth than AutoSAM and PerSAM, two existing prompt-based adaptation methods for SAM. 

\vspace*{-3pt}
\section{Conclusion}
\vspace*{-3pt}
This work proposes to automate a prompt-based universal model, such as MedSAM, by generating task-specific prompt embeddings directly from the image embedding of the foundation model. Our add-on module that can be integrated directly into MedSAM removes its dependence on user inputs. More importantly, by applying tightness and size constraints, our module can be trained effectively with only bounding box annotations while keeping MedSAM frozen. Furthermore, our 10-shot experiments has demonstrated that the number of samples required to train the model could be considerably reduced without a substantial degradation of the model performance. By adding a lightweight prompt module that can be trained with only few weak labels, MedSAM can efficiently be automated for specific tasks with minimal annotation hurdles.

\begin{credits}
\subsubsection{\discintname}
This work is supported by the Canada Research Chair on Shape Analysis in Medical Imaging, the Research Council of Canada (NSERC) and the Fonds de Recherche du Québec – Nature et Technologies (FRQNT).
\end{credits}
%
%
%
\bibliographystyle{splncs04}
\bibliography{mybibliography}
%

\end{document}